\documentclass[11pt,a4paper]{article}
\usepackage[hyperref]{ranlp2021}
\usepackage{times}
\usepackage{latexsym}
\usepackage{graphicx}
\usepackage{booktabs}
\graphicspath{ {./images/} }

\newcommand{\ra}[1]{\renewcommand{\arraystretch}{#1}}

\usepackage{microtype}

\aclfinalcopy 

\title{Can the Transformer Be Used as a Drop-in Replacement for RNNs in Text-Generating GANs?}

\author{Kevin Blin \\
  Section of Communication Systems \\
  Bachelor semester 6 \\ 
  EPFL, Lausanne, Switzerland \\
  \texttt{kevin.blin@epfl.ch} \\\And
  Andrei Kucharavy \\
  Distributed  Computing Laboratory\\
  EPFL, Lausanne, Switzerland \\
  \texttt{andrei.kucharavy@epfl.ch} \\}

\date{05.06.2021}

\begin{document}
\maketitle
\begin{abstract}
In this paper we address the problem of fine-tuned text generation with a limited computational budget. For that, we use a well-performing text generative adversarial network (GAN) architecture - Diversity-Promoting GAN (DPGAN), and attempted a drop-in replacement of the LSTM layer with a self-attention-based Transformer layer in order to leverage their efficiency. The resulting Self-Attention DPGAN (SADPGAN) was evaluated for performance, quality and diversity of generated text and stability. Computational experiments suggested that a transformer architecture is unable to drop-in replace the LSTM layer, under-performing during the pre-training phase and undergoing a complete mode collapse during the GAN tuning phase. Our results suggest that the transformer architecture need to be adapted before it can be used as a replacement for RNNs in text-generating GANs.

\end{abstract}



\section{Introduction}

Since the introduction of the Transformer in late 2017 by \citet{TheTransformerPaper}, pure self-attention architectures have become ubiquitous in the natural language processing community. 

Initially introduced by \citet{AttentionBengio2015}, the self-attention mechanism proved to be particularly useful in the context of neural machine translation to augment existing recurrent neural networks (RNNs). Recurrent neural networks (RNNs),  introduced by \citet{RNN_Hinton} and expanded by  \citet{LSTMOrigin, GRUOrigin}, were the state of the art in learning and generating sequential models, notably for texts. With the addition of the attention mechanism, in the context of machine translation, they could focus better on the words and word combinations corresponding to the same concepts in different languages and focus on learning equivalences between them as opposed to trying to infer them directly from the whole excerpts of parallel texts used for learning.

Following the initial introduction of the attention mechanism to augment the RNNs, architectures combining the two became de-facto state of the art. Stacking RNN layers, adding pass-through mechanisms and separating the architecture into encoder and decoder with an attention layer in the middle became a standard, powering among others Google's Neural Machine Translation system \citep{GoogleTranslateInDepth2016, GoogleTranslateOneShot2017}. Despite impressive performance, those architectures had a fundamental limitation to their ability to scale. The RNN training and evaluation are sequential by their nature, which means that architectures relying on the RNNs could hardly benefit from the arrival of massively parallel computing.

The innovation of the Transformer was to show that it was possible to learn sequence-to-sequence mapping while dispensing entirely with RNN layers, using only self-attention mechanisms ("Attention is all you need") - \citet{TheTransformerPaper}). By stacking several layers of self-attention networks to form an encoder and a decoder, as well as introducing multi-head architectures, where each layer of self-attention network could be trained in parallel, the Transformer and architectures derived from it scaled up easily and could be trained in parallel at a scale that was previously unreachable. 

That scalability enabled a continuous ramp-up of performance through parameter and training dataset size increase \citep{GPT3Paper2020}, which eventually hurt itself against the limitations of reasonable demands on computational power in deployments. In turn, this led to an extensive research into making Transformer-based architectures more efficient, focusing at first on specific instances \citep{TinyBERT, DistillBERT} and more recently on more general approaches (eg. \citet{WideButShortTransformers, NvidiaCompression, GoogleSwitchTransformers, MSEfficientTransformers}). Combined with the existence of specialized, energy-efficient hardware that is highly compatible with the Transformer architectures, this make Transformer-based architectures an attractive architecture to get the best value of limited computational budget.

The text-generating capabilities of the Transformer also gave rise to a new generation of models specialized in text generation. Rather than mapping texts between languages, they focused on mapping a prompt to a text that would follow it. Google's BERT model \citep{BERT2019} isolated and scaled up the Transformer encoder stack in order to perform masked training - predicting masked words in a text - referred to as \textit{autoencoding models} \citep{RNN_Hinton, Hinton2006AutoregressiveEncore, BengioAutoregressive2010Too}. On the other side of the spectrum, OpenAI's generative pre-training (GPT) family of models \cite{GPTPaper2018, GPT2Paper2019, GPT3Paper2020}, focused on approximating the decoder stack and using the tokens from the prompt in order to initialize the hidden state of the self-attention modules stack - referred to as \textit{autoregressive models} (by similarity with statistical autoregressive methods \citep{yule1926Autoregressive, wold1938Autoregressive, slutzky1937Autoregressive, boxJenkins1970Autoregressive}, see \citet{BengioAutoRegression} for machine learning applications).

Despite their impressive performance, when it comes to the text generation, both types of models are essentially autoregressive and are trained by max likelihood methods with regards to the training datasets. This poses several challenges. First, the autoregressive models learn the next token as a continuation of real texts they encounter in training, yet during the generation phase they continue from the text they themselves have generated. This means that during the generation phase they can rapidly go off the deeper end into an uncharted territory, they don't have a statistical model, and start generating degenerate output - a problem referred to as \textit{exposure biais} \citep{GenerativeModelsDegeneration2020}. So far, solutions to this problem, such as scheduled sampling \citep{ScheduledSamplingBengio2015}, are far from perfect and result in less diverse sampling and mode collapse \citep{ScheduledSamplingModeCollapse2015}. On top of that, the autoregressive nature of the model means that even in the territory where it has learned an appropriate token distribution, it will still be learning potentially undesirable biases \citep{DisabilityEmotionalBias2020}, with no means to correct them other than to curate the entire dataset used to train the model (reviewed in depth by \citet{StochasticParrotsGebru2021}).

Trying to learn the explicit statistical structure of natural language is, however, not the only way to train generative models. Adversarial Generative Networks (GANs) are a different training mode, where a generative model learns to generate outputs that are indistinguishable from the ones in the training dataset through a competition with a critic model, trained in tandem with it. Introduced by  \citep{GoodfellowGANs2014}, they are more robust to output degeneration, given that they always train in the generative mode, and require less computational resources than traditional auto regressive models. Besides, a number of different pre-trained critics can be used to eliminate undesired biases or on the contrary, introduce desired ones, such as specifying an artistic style of an image \citep{ArtisticStyleTransfert2015}.

The adversarial learning approach has been highly successful for training image generation models, allowing high-quality image generation \citep{BigGAN2019}, day-to-night or summer-to-winter image translation \citep{DayToNightConversionGAN2017}, or sketch-to-image translation \citep{SketchToImage2018}. GANs application to text generation, however, remained relatively limited. A major reason for that is that the sampling step leading to discrete and sequential token organization, needed for text generation, is problematic for gradient estimation, which is essential for training GANs. As such, most existing text-generating GANs rely on a max-likelihood autoregressive pre-training with the actual adversarial training phase being short and using a small learning rate, similar to a fine-tuning. Unfortunately such an approach failed to address the shortcomings of the purely autoregressive models they acquire during the pre-training phase. 

However, recently \citet{ScratchGAN2019} was able to demonstrate that it was possible to train a text-generating GAN architecture from scratch, thus avoiding entirely the problems encountered by autoregressive max-likelihood methods. A curious property of the Scratch-GAN he developed, is that while it seems to solve problems that plagues both RNNs and pure self-attention autoregressive models learning the explicit text token distribution, the authors still opted for RNN blocks, foregoing the advantages of Transformer architectures in NLP and self-attention in image-generating GANs \citep{SelfAttentionGANGoodfellow2019}.

Given the massive advantages of the Transformer presents when it comes to training as well as with regards to the amount of research performed to make them more efficient, as well as better capabilities of Transformers compared to GANs, we wanted to know if it was possible to perform a drop-in replacement of RNNs with Transformers. Such a Transformer-based GAN could provide two main benefits: the ability of produce higher quality samples at a reduced computational cost compared to traditional RNN-based GANs, and a more scalable language GAN.

In order to approach this question, we chose to perform an experimental evaluation, based on a classical text-generating GAN - Diversity Promoting GAN (DPGAN), developed by \citet{xu-etal-2018-diversity}. We chose DPGAN due to its straight-forwards architecture and training mode, similarity of its rewards structure to the state of the art text-generating GANs \citet{ScratchGAN2019} and the presence of the maximum likelihood pre-training, that we were expecting to be particularly favorable to the Transformer layers. We refer to the DPGAN with RNN layers replaced by a transformer as Self-attention DPGAN (SADPGAN).
\section{Related Work}

The power of the Transformer architecture in the language modelling tasks and its potential to further improve existing GAN architectures did not escape the attention of the machine learning community. So far, self-attention architectures in GANs focused on image generation tasks. Perhaps the two best known examples are TransGAN \citep{TransGAN} and Self-Attention GAN (SAGAN) \citep{SelfAttentionGANGoodfellow2019}. More recent advances, such as the introduction of Generative Adversarial Transformers by \citet{GenerativeAdversarialTransformers2021} have build on the Transformer architecture even further, enabling long-range correlation to improve over the existing state of art, showing a potential for it in the GAN setting. 
However, a common point between all of them is that they focus on the generation of images, rather than texts, and use only a single encoding layer for self attention or attempt to modify self-attention mechanism to better suite the image generation, diverging from the Transformer architecture. 

The approach to text generation by combining the Transformer and GANs that comes closest to ours is the SALSA-Text, developed by \citet{SalsaText2019}. SALSA-Text is a text-generating GAN build around a Transformer, discarding the original layer normalization and replacing it with the spectral layer normalization. In addition to that, SALSA-Text uses a modified Transformer architecture, with less layers and a different structure, as well as a specific training regimen, meaning it is more of a GAN built around a Transformer rather than a GAN where an RNN layer has been replaced with a pure self-attention based layer.

Here, we examine the Transformer applicability as a general-purpose element that can be drop-in in architectures requiring a latent space encoding and thus directly replace RNNs structures as LSTMs and Gated Recurrent Units (GRUs).

\section{Contribution and Outline}
Our contribution consisted in assessing the potential of Transformers as a drop-in replacement of LSTMs in text-generating GANs. Through three different experiment involving SADPGAN, we showed that Transformers, despite achieving remarkable results in several NLP tasks, fail to adapt to the adversarial learning and adversarial fine-tuning context \citep{AdversarialFineTuning}, causing SADPGAN to consistently come short of the DPGAN performance and to exhibit severe mode collapse. This results suggest that as of now, the Transformer can't be directly used as a replacement for LSTM without further architectural and training mode changes.

\section{Methodology}
Our work is built upon an existing PyTorch implementation of DPGAN by \citet{liu2020catgan}.

We used an iterative approach, first implementing a layer using exclusively Transformer encoders, then adding the decoder stack with masking and finally adding teacher forcing during training. For each of these steps we compared the training results to the original implementation.

Since our goal was to investigate the possibility of replacing LSTM layers in text-generating GANs by dropping in Transformer layers, and not to achieve new state of the art for text-generating GANs, we kept our models relatively small:  embedding and hidden dimension of 32, Transformer encoder/decoder with two layers, each with 4 attention heads of size 64. 

The training loop for both GANs consisted in 120 iterations of MLE pre-training for the generator followed by another 120 epochs of adversarial training between the discriminator and generator.
To assess the performance of both architectures, we used two negative log likelihood metrics, $NLL_{gen}$ and $NLL_{div}$ which measure the quality and respectively diversity of the generated text and (self)BLEU scores when using real data. For the NLL metrics lower values correspond to better results while for BLEU higher scores are desirable. The code developed for this project is available from \url{https://github.com/TheBlueHawk/RANLP21-70}, specifically the RANLP-2021 release. For the ease of use, it was packaged integrated into the code from \citet{liu2020catgan}.

\section{Results and Discussion}
\subsection{Experiment \#1}
Following the example set by \citet{TransGAN} in image GANs, we first tried to drop-in a Transformer encoder block, ensuring that it would properly fit the rest of the GAN architecture. 
For this preliminary experiment we trained both SADPGAN and DPGAN  on a synthetic dataset with a vocabulary of 5000 words. 

\begin{figure}[ht]
\caption{$NLL_{div}$ and $NLL_{gen}$ losses of SADPGAN and DPGAN during the first experiment.}
\centering
\includegraphics[width=0.4\textwidth]{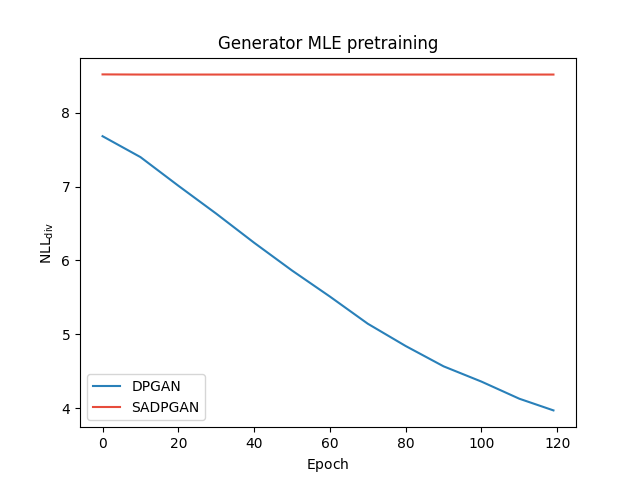}
\includegraphics[width=0.4\textwidth]{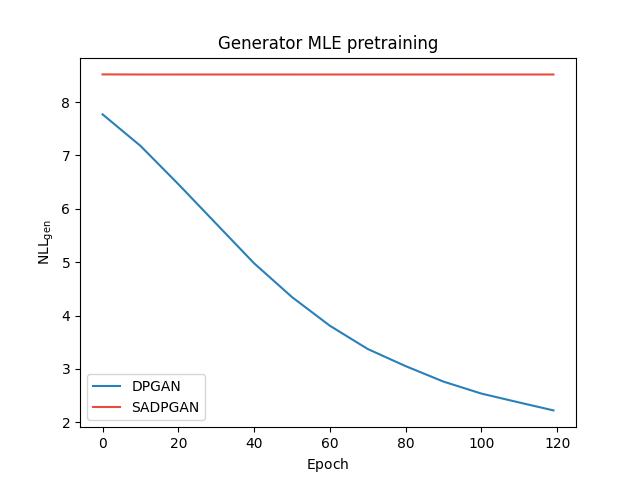}
\label{fig:NLL1}
\end{figure}

As shown in figure \ref{fig:NLL1}, for SADPGAN we didn't observe any improvement during MLE training. This can be explained by the choice of architecture: using only Transformer encoders without combining them with up-sampling layers for image generation or using bidirectional attention heads as in \citet{BERT2019}, doesn't allow the model to correctly learn how to produce realistic samples. 

\subsection{Experiment \#2}
Following the results of the first experiment, we modified the Transformer block by adding a Transformer decoder after the Transformer encoder. Contrary to an encoder-only architecture, the addition of the decoder require a second input, the output vector (also know as target), which is fed to the first masked multi-head attention layer of the decoder. 
To leave a maximum freedom to the decoder when generating sentences, we decided to feed it with an empty vector.
For this second experiment we trained both model on real data, specifically the \citet{COCOdataset} annotation dataset using a pre-trained Word2Vec embedding. Word samples were obtained in parallel using multinomial sampling. 

The set of modifications permitted SADPGAN to improve the quality of the generated text during the MLE pre-training iterations but the learning curve flattened out rather quickly and performed worse than the original implementation, with also a decrease of output diversity. 
During the adversarial training we observed a severe mode collapse: the original DPGAN produced very similar sentences while SADPGAN eventually produced exclusively empty sentences. 
See table \ref{tab:samples} for some samples randomly drawn at the end of pre-training and adversarial training.

\begin{table*}[ht]
    \centering
    \begin{tabular}{@{}rl@{}}
    \toprule
    DPGAN &\\
         pre & a person sits on black motorcycle on a busy bench near the side \\
        adv & a man riding a motorcycle down a street \\
         SADPGAN \\
         pre & flies a various a trash a while begs area in refrigerator \\
         adv & "empty sentence" \\
        \bottomrule
    \end{tabular}
    \caption{Text samples produced after pre-training (pre) and adversarial training (adv) by GAN architectures 
    }
    \label{tab:samples}
\end{table*}

The reason for this particular behaviour lies in the role of the target vector, which act as a ground truth in a teacher forcing scenario. This means that parallel sampling cannot be performed without forcing the use of the empty sentence as a reference. 

\begin{figure}[ht]
\caption{$NLL_{div}$ and $NLL_{gen}$ losses of SADPGAN and DPGAN during the second experiment.}
\centering
\includegraphics[width=0.4\textwidth]{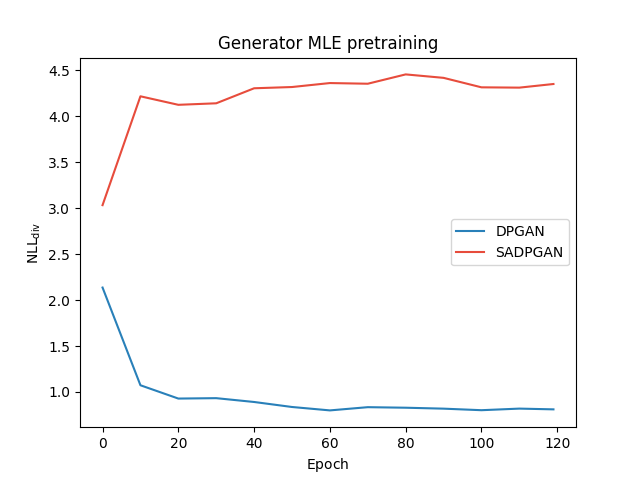}
\includegraphics[width=0.4\textwidth]{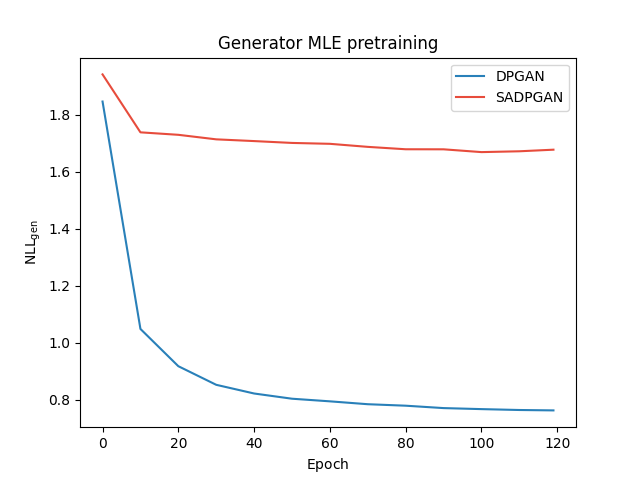}
\label{fig:NLL2}
\end{figure}

\subsection{Experiment \#3}
To address the problems of experiment \#2 we used  two modifications: a new target vector and auto-regressive sampling. As a target vector we used the input sentence to the encoder but shifted right by one position: this allowed a correct implementation of teacher forcing during the training phase. During the evaluation phase or when sampling, we don't want the teacher to influence the output of the block, and thus we used a word-by-word auto-regressive structure where the previously generated text is reused as the input of the target vector of the Transformer decoder.

The results of pre-training show a greater improvement of the quality of samples but at the cost of severe lack of diversity, resulting in common words repeated in sequence. Again, the adversarial training further exacerbate the issues resulting in total mode collapse. BLEU and self-BLEU scores for different n-grams at the end of pre-training and adversarial training are reported in table \ref{tab:bleu} for both DPGAN and SADPGAN.

\begin{table*}\centering
\ra{1}
\begin{tabular}{@{}rrrrrcrrrc@{}}\toprule
& \multicolumn{4}{c}{BLEU} & \phantom{abc}& \multicolumn{3}{c}{Self-BLEU} & \phantom{abc}\\
\cmidrule{2-5} \cmidrule{7-9}
& $n=2$ & $n=3$ & $n=4$ & $n=5$ && $n=2$ & $n=3$ & $n=4$\\ \midrule
DPGAN\\
pre & 0.73 & 0.497 & 0.306 & 0.185 && 0.77 & 0.537 & 0.327\\
adv & 0.845 & 0.737 & 0.578 & 0.455 && 0.993 & 0.989 & 0.982\\
SADPGAN\\
pre & 0.276 & 0.065 & 0.03 & 0.019 && 0.97 & 0.945 & 0.912\\
adv & \multicolumn{4}{c}{nan} & \phantom{abc}& \multicolumn{3}{c}{nan} & \phantom{abc} \\
\bottomrule
\end{tabular}
\caption{n-gram BLEU and self-BLEU scores for DPGAN and SADPGAN after pre-training and adversarial training}
\label{tab:bleu}
\end{table*}

\begin{figure}[ht]
\caption{$NLL_{div}$ and $NLL_{gen}$ losses of SADPGAN and DPGAN during the third experiment}
\centering
\includegraphics[width=0.4\textwidth]{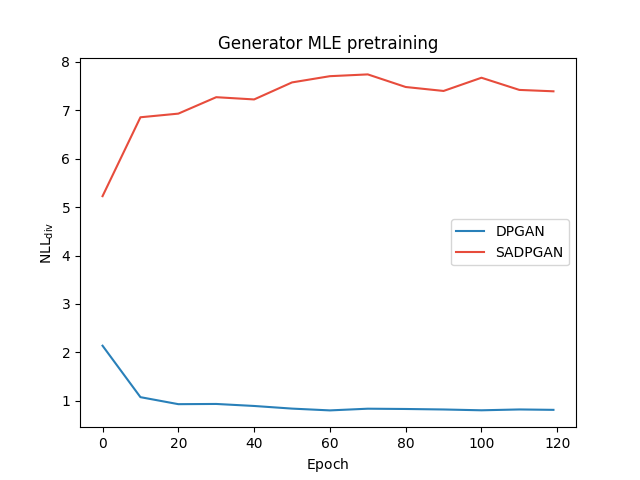}
\includegraphics[width=0.4\textwidth]{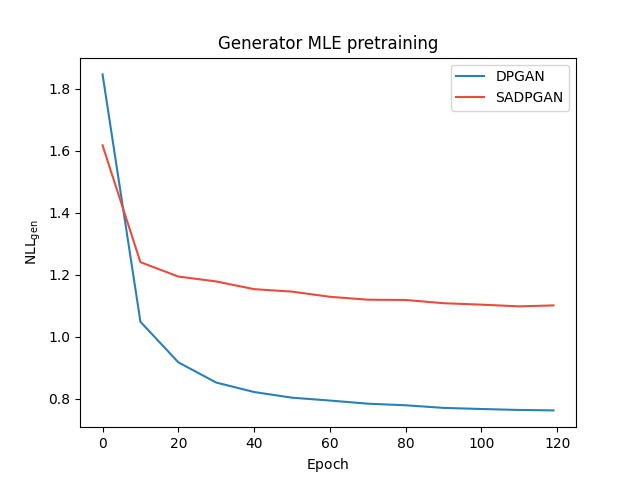}
\label{fig:NLL3}
\end{figure}

\section{Conclusion}

Following our results, we have observed that a Transformer architecture cannot be used as a simple drop-in replacement for RNNs in the context of text-generating GANs, at least not in its unmodified form. 

This result is not entirely surprising. Despite a great performance of the Transformer and its derivatives in different NLP tasks, Transformer-based architectures are all but simple to train, as reviewed in \citet{ProblemsWithTrainingTransformers2020}. 

Another problem with Transformer architectures is their tendency to overfit the distribution of the text they are learning and fail to generate novel text in case insufficiently diverse and varied training datasets are used. In the case of the DPGAN, the small size of the training dataset and the relatively small size of batches likely place us directly into the overfitting territory. Upon the transition into the adversarial training, this overfitting is likely mutually amplified, leading to more and more degenerate text outputs.

A more fundamental problem is that in DPGAN, we are still faced with the transition from the max-likelihood training regime to a generative regime. Which means that even with improved datasets, the generator is likely to wander off into the uncharted territory as it tries to generate a new token based on the tokens it has already generated rather than tokes sampled from the training dataset. While this problem is present as well with the RNNs, it seems that the output quality decay occurs faster with the Transformer-based architecture.

Despite the existing wealth of text-generating GANs, none, except for SALSA-text \citep{SalsaText2019} are Transformer-based, but persistently use RNNs, notably LSTM layers, including in the most recent, state-of-the-art ones, such as \citep{ScratchGAN2019}. In case of SALSA-text, a modified architecture and a different regularization methods are used, both in two specific setups. It seems that part of this trend in using RNNs rather than a Transformer-based architecture is rooted in the existence of fundamental differences in the way the Transformer learns compared to the RNNs, that make it particularly vulnerable to the hazards of adversarial training regiments.

As a result, we expect that developing text GAN architectures using self-attention based architectures instead of RNNs ones would require designing new GAN architecture from scratch to ensure a initialization and evaluation/reward structure that would be compatible with the the Transformer layer. A pure self-attention layer capable to be a drop-in replacement for RNNs is still to be developed.  Given the similarity between the adversarial phase occurring in the majority of current text-generating GANs and the adversarial fine-tuning mechanism (reviewed by eg. \citet{AdversarialFineTuning}), we expect that text-generating model fine-tuning to avoid undesirable patterns and adversarial prompts (such as exemplified in \citet{StochasticParrotsGebru2021}) would not be straightforwards and rely on the development of more robust and stable self-attention architectures.

\section{Future Work}

While this paper presents a negative result, we did not evaluate a number of approaches that could stabilize the training of pure self-attention architectures. Such approaches could prove to be key to the development of self-attention architectures that can be used as drop-in RNN layers replacements. 

An interesting avenue is to build upon the achievements of \citet{SalsaText2019}, and start with a reduced Transformer architecture, combined with a spectral normalization. Building up on this idea, Noise Stability Regularization proposed by \citet{NoiseStabilityRegularization2021}, suggested that noise regularization methods were capable to significantly improve the stability pre-trained pure self-attention generative networks fine-tuning, which could indicate an overall improvement in stability that would be visible in generative adversarial training. The regularization approach has been as well highlighted by several other publications, such as \citet{PostPreNormTransformers2019} and would be the first line of research to be investigated.

Another angle of attack would be to increase the amount of tokens used for initialization of the Transformer on the generator side as well as to perform the initialization on multiple levels. This approach have been shown to perform well in \citet{Autoprompt2020}. Similarly, existing Transformer architectures are known to require learning rate schedulers for training, something that seems to be entirely absent from almost all existing text-generating GAN architectures. It is possible that a learning rate scheduler needs to be incorporated into a Transformer layer for it to be able to become generally applicable. The importance of the learning rate scheduler for Transformers has been extensively documented in the past, notably by \citet{PopelTransformerTips2018} and could also be key in stabilizing the training of Transformer-based architectures.

Overall, in the absence of proof that pure self-attention architectures are inherently unstable in the adversarial training context, there is a number of potential approaches to make them work in the context of GANs and leverage their advantages that are to be explored.

\section*{Acknowledgments}

We would like to thank the Armasuisse - Swiss Cyber Defence Campus for the Distinguished Post Doctoral Fellowship supporting AK, and for providing the computational infrastructure for experiments, as well as Drs Rachid Guerraoui (EPFL) and Ljiljana Dolamic (CYD/Armasuisse), for oversight and general advice on this project, and finally Fabien Salvi (EPFL) for the technical support regarding the computational infrastructure organization.

\bibliographystyle{acl_natbib}
\bibliography{anthology,ranlp2021}

\begin{thebibliography}{48}
\expandafter\ifx\csname natexlab\endcsname\relax\def\natexlab#1{#1}\fi

\bibitem[{Bahdanau et~al.(2015)Bahdanau, Cho, and Bengio}]{AttentionBengio2015}
Dzmitry Bahdanau, Kyunghyun Cho, and Yoshua Bengio. 2015.
\newblock \href {http://arxiv.org/abs/1409.0473} {Neural machine translation by
  jointly learning to align and translate}.
\newblock In \emph{3rd International Conference on Learning Representations,
  {ICLR} 2015, San Diego, CA, USA, May 7-9, 2015, Conference Track
  Proceedings}.

\bibitem[{Bender et~al.(2021)Bender, Gebru, McMillan{-}Major, and
  Shmitchell}]{StochasticParrotsGebru2021}
Emily~M. Bender, Timnit Gebru, Angelina McMillan{-}Major, and Shmargaret
  Shmitchell. 2021.
\newblock \href {https://doi.org/10.1145/3442188.3445922} {On the dangers of
  stochastic parrots: Can language models be too big?}
\newblock In \emph{FAccT '21: 2021 {ACM} Conference on Fairness,
  Accountability, and Transparency, Virtual Event / Toronto, Canada, March
  3-10, 2021}, pages 610--623. {ACM}.

\bibitem[{Bengio et~al.(2015)Bengio, Vinyals, Jaitly, and
  Shazeer}]{ScheduledSamplingBengio2015}
Samy Bengio, Oriol Vinyals, Navdeep Jaitly, and Noam Shazeer. 2015.
\newblock \href
  {https://proceedings.neurips.cc/paper/2015/hash/e995f98d56967d946471af29d7bf99f1-Abstract.html}
  {Scheduled sampling for sequence prediction with recurrent neural networks}.
\newblock In \emph{Advances in Neural Information Processing Systems 28: Annual
  Conference on Neural Information Processing Systems 2015, December 7-12,
  2015, Montreal, Quebec, Canada}, pages 1171--1179.

\bibitem[{Bowman et~al.(2016)Bowman, Vilnis, Vinyals, Dai, J{\'{o}}zefowicz,
  and Bengio}]{BengioAutoRegression}
Samuel~R. Bowman, Luke Vilnis, Oriol Vinyals, Andrew~M. Dai, Rafal
  J{\'{o}}zefowicz, and Samy Bengio. 2016.
\newblock \href {https://doi.org/10.18653/v1/k16-1002} {Generating sentences
  from a continuous space}.
\newblock In \emph{Proceedings of the 20th {SIGNLL} Conference on Computational
  Natural Language Learning, CoNLL 2016, Berlin, Germany, August 11-12, 2016},
  pages 10--21. {ACL}.

\bibitem[{Box and Jenkins(1970)}]{boxJenkins1970Autoregressive}
George~EP Box and Gwilym~M Jenkins. 1970.
\newblock \emph{Time series analysis: forecasting and control}.
\newblock Holden-Day.

\bibitem[{Brock et~al.(2019)Brock, Donahue, and Simonyan}]{BigGAN2019}
Andrew Brock, Jeff Donahue, and Karen Simonyan. 2019.
\newblock \href {https://openreview.net/forum?id=B1xsqj09Fm} {Large scale {GAN}
  training for high fidelity natural image synthesis}.
\newblock In \emph{7th International Conference on Learning Representations,
  {ICLR} 2019, New Orleans, LA, USA, May 6-9, 2019}. OpenReview.net.

\bibitem[{Brown et~al.(2020)Brown, Mann, Ryder, Subbiah, Kaplan, Dhariwal,
  Neelakantan, Shyam, Sastry, Askell, Agarwal, Herbert{-}Voss, Krueger,
  Henighan, Child, Ramesh, Ziegler, Wu, Winter, Hesse, Chen, Sigler, Litwin,
  Gray, Chess, Clark, Berner, McCandlish, Radford, Sutskever, and
  Amodei}]{GPT3Paper2020}
Tom~B. Brown, Benjamin Mann, Nick Ryder, Melanie Subbiah, Jared Kaplan,
  Prafulla Dhariwal, Arvind Neelakantan, Pranav Shyam, Girish Sastry, Amanda
  Askell, Sandhini Agarwal, Ariel Herbert{-}Voss, Gretchen Krueger, Tom
  Henighan, Rewon Child, Aditya Ramesh, Daniel~M. Ziegler, Jeffrey Wu, Clemens
  Winter, Christopher Hesse, Mark Chen, Eric Sigler, Mateusz Litwin, Scott
  Gray, Benjamin Chess, Jack Clark, Christopher Berner, Sam McCandlish, Alec
  Radford, Ilya Sutskever, and Dario Amodei. 2020.
\newblock \href
  {https://proceedings.neurips.cc/paper/2020/hash/1457c0d6bfcb4967418bfb8ac142f64a-Abstract.html}
  {Language models are few-shot learners}.
\newblock In \emph{Advances in Neural Information Processing Systems 33: Annual
  Conference on Neural Information Processing Systems 2020, NeurIPS 2020,
  December 6-12, 2020, virtual}.

\bibitem[{Cho et~al.(2014)Cho, van Merrienboer, G{\"{u}}l{\c{c}}ehre, Bahdanau,
  Bougares, Schwenk, and Bengio}]{GRUOrigin}
Kyunghyun Cho, Bart van Merrienboer, {\c{C}}aglar G{\"{u}}l{\c{c}}ehre, Dzmitry
  Bahdanau, Fethi Bougares, Holger Schwenk, and Yoshua Bengio. 2014.
\newblock \href {https://doi.org/10.3115/v1/d14-1179} {Learning phrase
  representations using {RNN} encoder-decoder for statistical machine
  translation}.
\newblock In \emph{Proceedings of the 2014 Conference on Empirical Methods in
  Natural Language Processing, {EMNLP} 2014, October 25-29, 2014, Doha, Qatar,
  {A} meeting of SIGDAT, a Special Interest Group of the {ACL}}, pages
  1724--1734. {ACL}.

\bibitem[{Devlin et~al.(2019)Devlin, Chang, Lee, and Toutanova}]{BERT2019}
Jacob Devlin, Ming{-}Wei Chang, Kenton Lee, and Kristina Toutanova. 2019.
\newblock \href {https://doi.org/10.18653/v1/n19-1423} {{BERT:} pre-training of
  deep bidirectional transformers for language understanding}.
\newblock In \emph{Proceedings of the 2019 Conference of the North American
  Chapter of the Association for Computational Linguistics: Human Language
  Technologies, {NAACL-HLT} 2019, Minneapolis, MN, USA, June 2-7, 2019, Volume
  1 (Long and Short Papers)}, pages 4171--4186. Association for Computational
  Linguistics.

\bibitem[{Erhan et~al.(2010)Erhan, Bengio, Courville, Manzagol, Vincent, and
  Bengio}]{BengioAutoregressive2010Too}
Dumitru Erhan, Yoshua Bengio, Aaron~C. Courville, Pierre{-}Antoine Manzagol,
  Pascal Vincent, and Samy Bengio. 2010.
\newblock \href {https://dl.acm.org/citation.cfm?id=1756025} {Why does
  unsupervised pre-training help deep learning?}
\newblock \emph{J. Mach. Learn. Res.}, 11:625--660.

\bibitem[{Fedus et~al.(2021)Fedus, Zoph, and
  Shazeer}]{GoogleSwitchTransformers}
William Fedus, Barret Zoph, and Noam Shazeer. 2021.
\newblock \href {http://arxiv.org/abs/2101.03961} {Switch transformers: Scaling
  to trillion parameter models with simple and efficient sparsity}.
\newblock \emph{CoRR}, abs/2101.03961.

\bibitem[{Gagnon{-}Marchand et~al.(2019)Gagnon{-}Marchand, Sadeghi, Haidar, and
  Rezagholizadeh}]{SalsaText2019}
Jules Gagnon{-}Marchand, Hamed Sadeghi, Md.~Akmal Haidar, and Mehdi
  Rezagholizadeh. 2019.
\newblock \href {https://doi.org/10.1007/978-3-030-18305-9\_10} {{SALSA-TEXT:}
  self attentive latent space based adversarial text generation}.
\newblock In \emph{Advances in Artificial Intelligence - 32nd Canadian
  Conference on Artificial Intelligence, Canadian {AI} 2019, Kingston, ON,
  Canada, May 28-31, 2019, Proceedings}, volume 11489 of \emph{Lecture Notes in
  Computer Science}, pages 119--131. Springer.

\bibitem[{Gatys et~al.(2015)Gatys, Ecker, and
  Bethge}]{ArtisticStyleTransfert2015}
Leon~A. Gatys, Alexander~S. Ecker, and Matthias Bethge. 2015.
\newblock \href {http://arxiv.org/abs/1508.06576} {A neural algorithm of
  artistic style}.
\newblock \emph{CoRR}, abs/1508.06576.

\bibitem[{Goodfellow et~al.(2014)Goodfellow, Pouget{-}Abadie, Mirza, Xu,
  Warde{-}Farley, Ozair, Courville, and Bengio}]{GoodfellowGANs2014}
Ian~J. Goodfellow, Jean Pouget{-}Abadie, Mehdi Mirza, Bing Xu, David
  Warde{-}Farley, Sherjil Ozair, Aaron~C. Courville, and Yoshua Bengio. 2014.
\newblock \href
  {https://proceedings.neurips.cc/paper/2014/hash/5ca3e9b122f61f8f06494c97b1afccf3-Abstract.html}
  {Generative adversarial nets}.
\newblock In \emph{Advances in Neural Information Processing Systems 27: Annual
  Conference on Neural Information Processing Systems 2014, December 8-13 2014,
  Montreal, Quebec, Canada}, pages 2672--2680.

\bibitem[{Hinton et~al.(2006)Hinton, Osindero, and
  Teh}]{Hinton2006AutoregressiveEncore}
Geoffrey~E. Hinton, Simon Osindero, and Yee~Whye Teh. 2006.
\newblock \href {https://doi.org/10.1162/neco.2006.18.7.1527} {A fast learning
  algorithm for deep belief nets}.
\newblock \emph{Neural Comput.}, 18(7):1527--1554.

\bibitem[{Hochreiter and Schmidhuber(1997)}]{LSTMOrigin}
Sepp Hochreiter and J{\"{u}}rgen Schmidhuber. 1997.
\newblock \href {https://doi.org/10.1162/neco.1997.9.8.1735} {Long short-term
  memory}.
\newblock \emph{Neural Comput.}, 9(8):1735--1780.

\bibitem[{Holtzman et~al.(2020)Holtzman, Buys, Du, Forbes, and
  Choi}]{GenerativeModelsDegeneration2020}
Ari Holtzman, Jan Buys, Li~Du, Maxwell Forbes, and Yejin Choi. 2020.
\newblock \href {https://openreview.net/forum?id=rygGQyrFvH} {The curious case
  of neural text degeneration}.
\newblock In \emph{8th International Conference on Learning Representations,
  {ICLR} 2020, Addis Ababa, Ethiopia, April 26-30, 2020}. OpenReview.net.

\bibitem[{Hua et~al.(2021)Hua, Li, Dou, Xu, and
  Luo}]{NoiseStabilityRegularization2021}
Hang Hua, Xingjian Li, Dejing Dou, Cheng{-}Zhong Xu, and Jiebo Luo. 2021.
\newblock \href {https://www.aclweb.org/anthology/2021.naacl-main.258/} {Noise
  stability regularization for improving {BERT} fine-tuning}.
\newblock In \emph{Proceedings of the 2021 Conference of the North American
  Chapter of the Association for Computational Linguistics: Human Language
  Technologies, {NAACL-HLT} 2021, Online, June 6-11, 2021}, pages 3229--3241.
  Association for Computational Linguistics.

\bibitem[{Hudson and Zitnick(2021)}]{GenerativeAdversarialTransformers2021}
Drew~A. Hudson and C.~Lawrence Zitnick. 2021.
\newblock \href {http://arxiv.org/abs/2103.01209} {Generative adversarial
  transformers}.
\newblock \emph{CoRR}, abs/2103.01209.

\bibitem[{Huszar(2015)}]{ScheduledSamplingModeCollapse2015}
Ferenc Huszar. 2015.
\newblock \href {http://arxiv.org/abs/1511.05101} {How (not) to train your
  generative model: Scheduled sampling, likelihood, adversary?}
\newblock \emph{CoRR}, abs/1511.05101.

\bibitem[{Hutchinson et~al.(2020)Hutchinson, Prabhakaran, Denton, Webster,
  Zhong, and Denuyl}]{DisabilityEmotionalBias2020}
Ben Hutchinson, Vinodkumar Prabhakaran, Emily Denton, Kellie Webster, Yu~Zhong,
  and Stephen Denuyl. 2020.
\newblock \href {https://doi.org/10.18653/v1/2020.acl-main.487} {Social biases
  in {NLP} models as barriers for persons with disabilities}.
\newblock In \emph{Proceedings of the 58th Annual Meeting of the Association
  for Computational Linguistics, {ACL} 2020, Online, July 5-10, 2020}, pages
  5491--5501. Association for Computational Linguistics.

\bibitem[{Isola et~al.(2017)Isola, Zhu, Zhou, and
  Efros}]{DayToNightConversionGAN2017}
Phillip Isola, Jun{-}Yan Zhu, Tinghui Zhou, and Alexei~A. Efros. 2017.
\newblock \href {https://doi.org/10.1109/CVPR.2017.632} {Image-to-image
  translation with conditional adversarial networks}.
\newblock In \emph{2017 {IEEE} Conference on Computer Vision and Pattern
  Recognition, {CVPR} 2017, Honolulu, HI, USA, July 21-26, 2017}, pages
  5967--5976. {IEEE} Computer Society.

\bibitem[{Jeddi et~al.(2020)Jeddi, Shafiee, and Wong}]{AdversarialFineTuning}
Ahmadreza Jeddi, Mohammad~Javad Shafiee, and Alexander Wong. 2020.
\newblock \href {http://arxiv.org/abs/2012.13628} {A simple fine-tuning is all
  you need: Towards robust deep learning via adversarial fine-tuning}.
\newblock \emph{CoRR}, abs/2012.13628.

\bibitem[{Jiang et~al.(2021)Jiang, Chang, and Wang}]{TransGAN}
Yifan Jiang, Shiyu Chang, and Zhangyang Wang. 2021.
\newblock \href {http://arxiv.org/abs/2102.07074} {Transgan: Two transformers
  can make one strong {GAN}}.
\newblock \emph{CoRR}, abs/2102.07074.

\bibitem[{Jiao et~al.(2020)Jiao, Yin, Shang, Jiang, Chen, Li, Wang, and
  Liu}]{TinyBERT}
Xiaoqi Jiao, Yichun Yin, Lifeng Shang, Xin Jiang, Xiao Chen, Linlin Li, Fang
  Wang, and Qun Liu. 2020.
\newblock \href {https://doi.org/10.18653/v1/2020.findings-emnlp.372}
  {Tinybert: Distilling {BERT} for natural language understanding}.
\newblock In \emph{Proceedings of the 2020 Conference on Empirical Methods in
  Natural Language Processing: Findings, {EMNLP} 2020, Online Event, 16-20
  November 2020}, volume {EMNLP} 2020 of \emph{Findings of {ACL}}, pages
  4163--4174. Association for Computational Linguistics.

\bibitem[{Johnson et~al.(2017)Johnson, Schuster, Le, Krikun, Wu, Chen, Thorat,
  Vi{\'{e}}gas, Wattenberg, Corrado, Hughes, and
  Dean}]{GoogleTranslateOneShot2017}
Melvin Johnson, Mike Schuster, Quoc~V. Le, Maxim Krikun, Yonghui Wu, Zhifeng
  Chen, Nikhil Thorat, Fernanda~B. Vi{\'{e}}gas, Martin Wattenberg, Greg
  Corrado, Macduff Hughes, and Jeffrey Dean. 2017.
\newblock \href {https://transacl.org/ojs/index.php/tacl/article/view/1081}
  {Google's multilingual neural machine translation system: Enabling zero-shot
  translation}.
\newblock \emph{Trans. Assoc. Comput. Linguistics}, 5:339--351.

\bibitem[{Li et~al.(2020)Li, Wallace, Shen, Lin, Keutzer, Klein, and
  Gonzalez}]{WideButShortTransformers}
Zhuohan Li, Eric Wallace, Sheng Shen, Kevin Lin, Kurt Keutzer, Dan Klein, and
  Joseph~E. Gonzalez. 2020.
\newblock \href {http://arxiv.org/abs/2002.11794} {Train large, then compress:
  Rethinking model size for efficient training and inference of transformers}.
\newblock \emph{CoRR}, abs/2002.11794.

\bibitem[{Lin et~al.(2014)Lin, Maire, Belongie, Bourdev, Girshick, Hays,
  Perona, Ramanan, Doll{\'{a}}r, and Zitnick}]{COCOdataset}
Tsung{-}Yi Lin, Michael Maire, Serge~J. Belongie, Lubomir~D. Bourdev, Ross~B.
  Girshick, James Hays, Pietro Perona, Deva Ramanan, Piotr Doll{\'{a}}r, and
  C.~Lawrence Zitnick. 2014.
\newblock \href {http://arxiv.org/abs/1405.0312} {Microsoft {COCO:} common
  objects in context}.
\newblock \emph{CoRR}, abs/1405.0312.

\bibitem[{Liu et~al.(2020{\natexlab{a}})Liu, Liu, Gao, Chen, and
  Han}]{ProblemsWithTrainingTransformers2020}
Liyuan Liu, Xiaodong Liu, Jianfeng Gao, Weizhu Chen, and Jiawei Han.
  2020{\natexlab{a}}.
\newblock \href {https://doi.org/10.18653/v1/2020.emnlp-main.463}
  {Understanding the difficulty of training transformers}.
\newblock In \emph{Proceedings of the 2020 Conference on Empirical Methods in
  Natural Language Processing, {EMNLP} 2020, Online, November 16-20, 2020},
  pages 5747--5763. Association for Computational Linguistics.

\bibitem[{Liu et~al.(2020{\natexlab{b}})Liu, Wang, and Liang}]{liu2020catgan}
Zhiyue Liu, Jiahai Wang, and Zhiwei Liang. 2020{\natexlab{b}}.
\newblock Catgan: Category-aware generative adversarial networks with
  hierarchical evolutionary learning for category text generation.
\newblock In \emph{Proceedings of the AAAI Conference on Artificial
  Intelligence}.

\bibitem[{Lu et~al.(2018)Lu, Wu, Tai, and Tang}]{SketchToImage2018}
Yongyi Lu, Shangzhe Wu, Yu{-}Wing Tai, and Chi{-}Keung Tang. 2018.
\newblock \href {https://doi.org/10.1007/978-3-030-01270-0\_13} {Image
  generation from sketch constraint using contextual {GAN}}.
\newblock In \emph{Computer Vision - {ECCV} 2018 - 15th European Conference,
  Munich, Germany, September 8-14, 2018, Proceedings, Part {XVI}}, volume 11220
  of \emph{Lecture Notes in Computer Science}, pages 213--228. Springer.

\bibitem[{Mandava et~al.(2020)Mandava, Migacz, and Florea}]{NvidiaCompression}
Swetha Mandava, Szymon Migacz, and Alex~Fit Florea. 2020.
\newblock \href {http://arxiv.org/abs/2009.04534} {Pay attention when
  required}.
\newblock \emph{CoRR}, abs/2009.04534.

\bibitem[{de~Masson~d'Autume et~al.(2019)de~Masson~d'Autume, Mohamed, Rosca,
  and Rae}]{ScratchGAN2019}
Cyprien de~Masson~d'Autume, Shakir Mohamed, Mihaela Rosca, and Jack~W. Rae.
  2019.
\newblock \href
  {https://proceedings.neurips.cc/paper/2019/hash/a6ea8471c120fe8cc35a2954c9b9c595-Abstract.html}
  {Training language gans from scratch}.
\newblock In \emph{Advances in Neural Information Processing Systems 32: Annual
  Conference on Neural Information Processing Systems 2019, NeurIPS 2019,
  December 8-14, 2019, Vancouver, BC, Canada}, pages 4302--4313.

\bibitem[{Nguyen and Salazar(2019)}]{PostPreNormTransformers2019}
Toan~Q. Nguyen and Julian Salazar. 2019.
\newblock \href {http://arxiv.org/abs/1910.05895} {Transformers without tears:
  Improving the normalization of self-attention}.
\newblock \emph{CoRR}, abs/1910.05895.

\bibitem[{Popel and Bojar(2018)}]{PopelTransformerTips2018}
Martin Popel and Ondrej Bojar. 2018.
\newblock \href {http://ufal.mff.cuni.cz/pbml/110/art-popel-bojar.pdf}
  {Training tips for the transformer model}.
\newblock \emph{Prague Bull. Math. Linguistics}, 110:43--70.

\bibitem[{Radford et~al.(2018)Radford, Narasimhan, Salimans, and
  Sutskever}]{GPTPaper2018}
Alec Radford, Karthik Narasimhan, Tim Salimans, and Ilya Sutskever. 2018.
\newblock Improving language understanding by generative pre-training.

\bibitem[{Radford et~al.(2019)Radford, Wu, Child, Luan, Amodei, and
  Sutskever}]{GPT2Paper2019}
Alec Radford, Jeffrey Wu, Rewon Child, David Luan, Dario Amodei, and Ilya
  Sutskever. 2019.
\newblock Language models are unsupervised multitask learners.
\newblock \emph{OpenAI blog}, 1(8):9.

\bibitem[{Ren et~al.(2021)Ren, Rajbhandari, Aminabadi, Ruwase, Yang, Zhang, Li,
  and He}]{MSEfficientTransformers}
Jie Ren, Samyam Rajbhandari, Reza~Yazdani Aminabadi, Olatunji Ruwase, Shuangyan
  Yang, Minjia Zhang, Dong Li, and Yuxiong He. 2021.
\newblock \href {https://www.usenix.org/conference/atc21/presentation/ren-jie}
  {Zero-offload: Democratizing billion-scale model training}.
\newblock In \emph{2021 {USENIX} Annual Technical Conference, {USENIX} {ATC}
  2021, July 14-16, 2021}, pages 551--564. {USENIX} Association.

\bibitem[{Rumelhart et~al.(1986)Rumelhart, Hinton, and Williams}]{RNN_Hinton}
D.~E. Rumelhart, G.~E. Hinton, and R.~J. Williams. 1986.
\newblock \emph{Learning Internal Representations by Error Propagation}, page
  318–362. MIT Press, Cambridge, MA, USA.

\bibitem[{Sanh et~al.(2019)Sanh, Debut, Chaumond, and Wolf}]{DistillBERT}
Victor Sanh, Lysandre Debut, Julien Chaumond, and Thomas Wolf. 2019.
\newblock \href {http://arxiv.org/abs/1910.01108} {Distilbert, a distilled
  version of {BERT:} smaller, faster, cheaper and lighter}.
\newblock \emph{CoRR}, abs/1910.01108.

\bibitem[{Shin et~al.(2020)Shin, Razeghi, IV, Wallace, and
  Singh}]{Autoprompt2020}
Taylor Shin, Yasaman Razeghi, Robert L.~Logan IV, Eric Wallace, and Sameer
  Singh. 2020.
\newblock \href {https://doi.org/10.18653/v1/2020.emnlp-main.346} {Autoprompt:
  Eliciting knowledge from language models with automatically generated
  prompts}.
\newblock In \emph{Proceedings of the 2020 Conference on Empirical Methods in
  Natural Language Processing, {EMNLP} 2020, Online, November 16-20, 2020},
  pages 4222--4235. Association for Computational Linguistics.

\bibitem[{Slutzky(1937)}]{slutzky1937Autoregressive}
Eugen Slutzky. 1937.
\newblock The summation of random causes as the source of cyclic processes.
\newblock \emph{Econometrica: Journal of the Econometric Society}, pages
  105--146.

\bibitem[{Vaswani et~al.(2017)Vaswani, Shazeer, Parmar, Uszkoreit, Jones,
  Gomez, Kaiser, and Polosukhin}]{TheTransformerPaper}
Ashish Vaswani, Noam Shazeer, Niki Parmar, Jakob Uszkoreit, Llion Jones,
  Aidan~N. Gomez, Lukasz Kaiser, and Illia Polosukhin. 2017.
\newblock \href
  {https://proceedings.neurips.cc/paper/2017/hash/3f5ee243547dee91fbd053c1c4a845aa-Abstract.html}
  {Attention is all you need}.
\newblock In \emph{Advances in Neural Information Processing Systems 30: Annual
  Conference on Neural Information Processing Systems 2017, December 4-9, 2017,
  Long Beach, CA, {USA}}, pages 5998--6008.

\bibitem[{Wold(1938)}]{wold1938Autoregressive}
Herman Wold. 1938.
\newblock \emph{A study in the analysis of stationary time series}.
\newblock Ph.D. thesis, Almqvist \& Wiksell.

\bibitem[{Wu et~al.(2016)Wu, Schuster, Chen, Le, Norouzi, Macherey, Krikun,
  Cao, Gao, Macherey, Klingner, Shah, Johnson, Liu, Kaiser, Gouws, Kato, Kudo,
  Kazawa, Stevens, Kurian, Patil, Wang, Young, Smith, Riesa, Rudnick, Vinyals,
  Corrado, Hughes, and Dean}]{GoogleTranslateInDepth2016}
Yonghui Wu, Mike Schuster, Zhifeng Chen, Quoc~V. Le, Mohammad Norouzi, Wolfgang
  Macherey, Maxim Krikun, Yuan Cao, Qin Gao, Klaus Macherey, Jeff Klingner,
  Apurva Shah, Melvin Johnson, Xiaobing Liu, Lukasz Kaiser, Stephan Gouws,
  Yoshikiyo Kato, Taku Kudo, Hideto Kazawa, Keith Stevens, George Kurian,
  Nishant Patil, Wei Wang, Cliff Young, Jason Smith, Jason Riesa, Alex Rudnick,
  Oriol Vinyals, Greg Corrado, Macduff Hughes, and Jeffrey Dean. 2016.
\newblock \href {http://arxiv.org/abs/1609.08144} {Google's neural machine
  translation system: Bridging the gap between human and machine translation}.
\newblock \emph{CoRR}, abs/1609.08144.

\bibitem[{Xu et~al.(2018)Xu, Ren, Lin, and Sun}]{xu-etal-2018-diversity}
Jingjing Xu, Xuancheng Ren, Junyang Lin, and Xu~Sun. 2018.
\newblock \href {https://doi.org/10.18653/v1/D18-1428} {Diversity-promoting
  {GAN}: A cross-entropy based generative adversarial network for diversified
  text generation}.
\newblock In \emph{Proceedings of the 2018 Conference on Empirical Methods in
  Natural Language Processing}, pages 3940--3949, Brussels, Belgium.
  Association for Computational Linguistics.

\bibitem[{Yule(1926)}]{yule1926Autoregressive}
G~Udny Yule. 1926.
\newblock Why do we sometimes get nonsense-correlations between time-series?--a
  study in sampling and the nature of time-series.
\newblock \emph{Journal of the royal statistical society}, 89(1):1--63.

\bibitem[{Zhang et~al.(2019)Zhang, Goodfellow, Metaxas, and
  Odena}]{SelfAttentionGANGoodfellow2019}
Han Zhang, Ian~J. Goodfellow, Dimitris~N. Metaxas, and Augustus Odena. 2019.
\newblock \href {http://proceedings.mlr.press/v97/zhang19d.html}
  {Self-attention generative adversarial networks}.
\newblock In \emph{Proceedings of the 36th International Conference on Machine
  Learning, {ICML} 2019, 9-15 June 2019, Long Beach, California, {USA}},
  volume~97 of \emph{Proceedings of Machine Learning Research}, pages
  7354--7363. {PMLR}.

\end{thebibliography}

\end{document}